\documentclass[letterpaper]{article} %
\usepackage{aaai2026}  %

\usepackage{times}  
\usepackage{helvet}  
\usepackage{courier}  
\usepackage[hyphens]{url}  
\usepackage{graphicx} 
\usepackage{natbib}  
\usepackage{caption} 
\usepackage{algorithm}
\usepackage{algorithmic}
\usepackage{amsmath}
\usepackage{amssymb}
\usepackage{multirow}
\usepackage{booktabs}
\usepackage{newfloat}
\usepackage{listings}
\DeclareCaptionStyle{ruled}{labelfont=normalfont,labelsep=colon,strut=off} %
\floatstyle{ruled}
\newfloat{listing}{tb}{lst}{}
\floatname{listing}{Listing}
\title{Retrieval with Multiple Query Vectors through Anomalous Pattern Detection}

\author{
    Allassan Tchangmena A Nken\textsuperscript{\rm 1}\thanks{This work was done during an internship at IBM Research.},
    Baimam Boukar Jean Jacques\textsuperscript{\rm 2, *}, \\ 
    Miriam Rateike\textsuperscript{\rm 3, 4},
    Celia Cintas\textsuperscript{\rm 3},
    Skyler Speakman\textsuperscript{\rm 3}
}
\affiliations{
    \textsuperscript{\rm 1}IBM,\\
    \textsuperscript{\rm 2}University of Galway, 
    \textsuperscript{\rm 3}Carnegie Mellon University Africa, 
    \textsuperscript{\rm 4}University of Tübingen, \\ 
a.tchangmenaanken1@universityofgalway.ie, bbaimamb@andrew.cmu.edu, \\
miriam.rateike@ibm.com, celia.cintas@ibm.com, skyler@ke.ibm.com
}

\everypar{\looseness=-1}
\newcommand{\Appendix}{Appendix}
\newcommand{\Figure}{Figure}

\definecolor{bleudefrance}{rgb}{0.19, 0.55, 0.91}
\definecolor{asparagus}{rgb}{0.53, 0.66, 0.42}
\definecolor{chocolate}{rgb}{0.82, 0.41, 0.12}
\definecolor{cadmiumgreen}{rgb}{0.0, 0.42, 0.24}
\definecolor{ao(english)}{rgb}{0.0, 0.5, 0.0}
\definecolor{blue-violet}{rgb}{0.54, 0.17, 0.89}
\definecolor{bittersweet}{rgb}{1.0, 0.44, 0.37}

\definecolor{nightblue}{RGB}{25,25,112}
\definecolor{babyblueeyes}{rgb}{0.63, 0.79, 0.95}
\definecolor{bluebell}{rgb}{0.64, 0.64, 0.82}
\definecolor{bubblegum}{rgb}{0.99, 0.76, 0.8}
 	\definecolor{carnationpink}{rgb}{1.0, 0.65, 0.79}
\definecolor{bleudefrance}{rgb}{0.19, 0.55, 0.91}
\definecolor{blush}{rgb}{0.87, 0.36, 0.51}

\newcommand{\database}{D}
\newcommand{\queryset}{Q}
\newcommand{\vectdim}{L}
\newcommand{\classset}{C}
\newcommand{\setretrieved}{R}
\newcommand{\numdb}{N}
\newcommand{\numqueries}{M}
\newcommand{\numretrieved}{k}
\newcommand{\KDTpre}{KDTree pre}
\newcommand{\KDTpost}{KDTree post}
\newcommand{\KDTree}{KDTree}

\newcommand{\mnist}{MNIST}
\newcommand{\fashionmnist}{Fashion-MNIST}
\newcommand{\persona}{Persona}
\newcommand{\tabulardata}{Tabular}
\newcommand{\ailuminate}{Ailuminate}

\definecolor{bleudefrance}{rgb}{0.19, 0.55, 0.91}
\definecolor{asparagus}{rgb}{0.53, 0.66, 0.42}
\definecolor{chocolate}{rgb}{0.82, 0.41, 0.12}
\definecolor{cadmiumgreen}{rgb}{0.0, 0.42, 0.24}
\definecolor{ao(english)}{rgb}{0.0, 0.5, 0.0}
\definecolor{blue-violet}{rgb}{0.54, 0.17, 0.89}
\definecolor{bittersweet}{rgb}{1.0, 0.44, 0.37}

\definecolor{nightblue}{RGB}{25,25,112}
\definecolor{babyblueeyes}{rgb}{0.63, 0.79, 0.95}
\definecolor{bluebell}{rgb}{0.64, 0.64, 0.82}
\definecolor{bubblegum}{rgb}{0.99, 0.76, 0.8}
 	\definecolor{carnationpink}{rgb}{1.0, 0.65, 0.79}
\definecolor{bleudefrance}{rgb}{0.19, 0.55, 0.91}
\definecolor{blush}{rgb}{0.87, 0.36, 0.51}

\begin{document}

\maketitle

\begin{abstract}
A classical vector retrieval problem typically considers a \emph{single} query embedding vector
as input and retrieves the most similar embedding vectors from a vector database. However, complex reasoning and retrieval tasks frequently require \emph{multiple query vectors}, rather than a single one. In this work, we propose a
retrieval method that considers multiple query vectors simultaneously and retrieves the most relevant vectors from the database using concepts from anomalous pattern detection. Specifically, our approach leverages a set of query vectors $Q$ (with $|Q|\geq 1$), and identifies the subset of vector dimensions within $Q$ that standout (anomalous) from the rest of dimensions. Next, we scan the vector database to retrieve the set of vectors that are also anomalous across the previously identified vector dimensions and return them as our retrieved set of vectors. We validate our approach on two image datasets, a text dataset, and a tabular dataset. Overall, we observe that, across most datasets, larger query sets lead to improved retrieval performance. The improvement is most pronounced when increasing the query sets  from 1 to 8, while the gains become smaller beyond that. 
\end{abstract}
\section{Introduction}
\label{sec:introduction}

Information retrieval (IR) underpins a wide range of applications such as question answering \cite{karpukhin2020dense}, image search \cite{oquab2024dinov2}, and more recently has become central in retrieval-augmented generation (RAG) for large language models (LLMs), including retrieval as a reasoning action \cite{yao2022react},  retrieval for self-critique/self-verification \cite{asai2024self}, and  retrieval-in-the-loop during LLM training \cite{tang2024self}. In addition, an increasing number of companies are adopting IR systems as they migrate from traditional relational databases to vector databases\footnote{Vector databases store data points as high-dimensional numerical vectors.}, owing to their ability to efficiently handle low-latency queries.\footnote{\url{https://www.marketresearch.com/Global-Industry-Analysts-v1039/Vector-Databases-41409421/}, last accessed on 05.11.2025.} 

Most retrieval systems operate in a \emph{single-query} setting.
For example, assume a news retrieval scenario, where a user provides a query sentence about an event, which is encoded into a single vector representation, such as the final token embedding from a large language model (LLM).  
The retrieval system then returns articles from a database, which are relevant to this query vector.
In practice, however, retrieval tasks often involve \emph{multiple query vectors}.  
For instance, a user may instead provide a paragraph summarizing the event, where each sentence captures a different aspect (such as the location, participants, or outcomes) and is encoded as a separate query vector.
Existing retrieval methods typically address this challenge by either aggregating multiple query vectors into a single embedding via simple heuristics~\cite{li2016deep, wang2022multi}, or by treating each query independently~\cite{xu2016msr}.
Both lines of work then apply standard single-query retrieval methods {such as  cosine similarity~\cite{venkatesh2024enhancing}, inner product~\cite{li2017fexipro}, or Euclidean distance~\cite{hertz2004learning}, which assume that relevance corresponds to proximity in the vector space. In addition, these approaches discard potentially valuable inter-query vector relationships~\cite{liu2024multi}, which may yield more contextually relevant retrieval results.}

In practice, however, this assumption may fail, as learned embeddings often encode complex, nonlinear, and context-dependent relationships that fixed distance metrics fail to capture~\cite{zhou2022problems}, particularly in high-dimensional LLM representations.

In this work, we propose a novel retrieval approach that (i) supports retrieval with multiple query vectors and (ii) defines relevance through an \emph{anomaly scoring function}, rather than through traditional distance-based measures.  
Unlike many existing methods~\cite{cunningham2020k,douze2025faiss},  our approach does not require the number of vectors to be retrieved to be specified a priori.

Our key hypothesis is that the query vectors share an \emph{anomalous pattern}, and that the most relevant database vectors in $\database$ are those exhibiting the same pattern. 
We extend prior work on deep scanning for anomalous pattern detection~\cite{mcfowland2013fast, cintas2021detecting, cintas2022towards, rateike2023weakly} to the retrieval setting, introducing a two-step process: (i) given a set of query vectors, we identify a subset of vector dimensions that jointly form an anomalous pattern; and  
(ii) using these dimensions, we retrieve database vectors that exhibit a similar joint anomalous pattern.
We empirically validate our approach on LLM text embeddings, reduced-dimensional image embeddings, and tabular data. We  compare against two \KDTree\ baselines, which we adapted to the multi query vector setting.  
Our results show that, particularly for LLM text embeddings, our method consistently outperforms the baselines in both precision and recall.

\section{Retrieval through Anomalous \\ Pattern Detection}

\subsection{Problem Formulation}\label{problem}
Let $\database = \{d_{1}, d_{2}, \ldots, d_{\numdb}\}$ be a database of vectors and $\queryset = \{q_{1}, q_{2}, \ldots, q_{\numqueries}\}$ a set of query vectors, where each vector $d_i, q_j \in \mathbb{R}^\vectdim$ and $\numdb, \numqueries \geq 1$.
These vectors could be embedding vectors from an LLM, low-dimensional representations of image data, or numerical vectors from a tabular dataset. 
Vectors in $\database$ and $\queryset$ are assumed to be from the same vector space and as such have the same number of dimensions (length) $\vectdim$. 
Given a set of query vectors $\queryset$, the goal is to retrieve 
the most relevant subset of database vectors $\setretrieved = \{r_1, r_2, \ldots, r_{\numretrieved}\} \subseteq \database$.

\subsection{Scanning Algorithm for Retrieval}

 We hypothesize that the vectors in $\queryset$ share an \emph{anomalous pattern}, and that the most relevant vectors in the database $\database$ are those exhibiting the same pattern.  
We characterize such an anomalous pattern as a subset of vector dimensions where the corresponding values in $\queryset$ jointly significantly deviate from their distribution in $\database$.
Our approach builds on an scanning algorithm for anomalous pattern detection introduced by~\citet{mcfowland2013fast} and later extended to detect anomalous patterns in embedding vectors~\cite{cintas2021detecting, cintas2022towards, rateike2023weakly}. We extend this framework to the retrieval setting outlined above. 

Formally, for any subset of vectors $V_S \subseteq V$, where each vector $v \in \mathbb{R}^\vectdim$, and any subset of vector dimensions $O_S \subseteq [\vectdim] := \{1, \ldots, \vectdim\}$, we refer to a \emph{subset} as 
$S = V_S \times O_S = \{(v, l) \mid v \in V_S,\; l \in O_S\}$,
that is, a set of all vector–dimension pairs obtained by restricting the vectors in $V_S$ to the dimensions in $O_S$.
Let further $F(S) \to \mathbb{R}$ be a scoring function that assigns an anomaly (or relevance) score to each such subset $S$. 
Prior work \cite{rateike2023weakly,cintas2022towards} has sought to identify a subset $S$ that maximizes $F(S)$ by jointly optimizing over subsets of vectors $V_S$ and the subsets of dimensions $O_S$. 
In contrast, we propose a two-step approach. 
First, we fix the vector subset $V_S$ (corresponding to the query vector set $\queryset$) and identify the most informative subset of vector dimensions $O^*_S$. Then, we fix this subset of vector dimensions $O^*_S$ and retrieve the most relevant vectors $V_S$ (corresponding to subset of retrieved database vectors ${\setretrieved \in \database}$).

\paragraph{Step 1: Identify subset of vector dimensions given the set of query vectors.} 

First, we find the most relevant dimensions that best capture the distinguishing characteristics of the set of queries.
Formally, given the set of queries $Q$, we seek for the subset of dimensions $O^{\star}_S$ that maximizes the scoring function $F(S_{\text{Step1}})$, where  
$S_{\text{Step1}} = Q \times O^{\star}_S$,  
that is, the subset obtained by restricting the query vectors in $Q$ to the set of vector dimensions $O^{\star}_S$. 

\paragraph{Step 2: Retrieve subset of database vectors given the subset of vector dimensions.} 
Then, we retrieve the vectors that are most relevant according to the previously identified vector dimensions.
More formally, given the subset of vector dimensions from Step 1, $O^*_S$, we seek for the subset of database vectors $\setretrieved^{\star}_S \subseteq \database$ that maximize the scoring function $F(S_{\text{Step2}})$, where  
$S_{\text{Step2}} = \setretrieved^{\star}_S \times O_S$,  
that is, the subset obtained by restricting the query vectors in $\setretrieved$ to the set of vector dimensions $O^{\star}_S$.

\subsection{Finding Relevant Patterns in the Query Set}
Recall that we assume vectors in $\queryset$ share an anomalous
pattern, and that the most relevant vectors in the database
$\database$ are those exhibiting the same pattern.
 We characterize an anomalous pattern in $\queryset$ by a subset of vector dimensions where the corresponding values are particularly low compared to their distribution in $\database$.\footnote{An anomalous pattern may also be characterized by particularly high values, or a combination of both; we will address this in future work.}
We consider a vector's value along dimension $l$ to be \emph{low} if it lies in the lower $\alpha$-tail of the value distribution for that dimension. For a normal distribution, this corresponds to values significantly below the mean.

\paragraph{Computing $\mathbf{z}$-scores.}
To identify dimensions along which the query vectors in $\queryset$ exhibit significantly low values, we standardize each dimension using $z$-scores. 
For each query vector $m$ and dimension $l$, the $z$-score for a single query vector entry $q_{m,l}$ is
$z_{ml} = \frac{q_{ml} - \mu_l}{\sigma_l}$,
where $\mu_l$ and $\sigma_l$ are the mean and standard deviation of values along dimension $l$ computed over all vectors in $\database$.
We assume each vector dimension follows an approximately normal distribution, and further compute $\textsc{CDF}(z_{ml})$ for all $m$ and $l$, where $\textsc{CDF}$ is the cumulative distribution function of the standard normal distribution.

\paragraph{Scoring Function.}

Following prior work~\cite{cintas2021detecting, rateike2023weakly}, we use a non-parametric scan statistic to compute the anomaly (or relevance) score. The scoring function takes the general form:
\begin{equation}
F(S) = \max_{\alpha} \phi\big(\alpha, N_{\alpha}(S), N(S)\big),
\end{equation}
where $\alpha$ represents a significance level, $\phi$ is a suitable goodness-of-fit statistic (for instance, we use Berk–Jones~\cite{berk1979goodness}), $N(S)$ is the number of values in the subset $S$, and $N_{\alpha}(S)$ is the number of values within $S$ that are below the threshold $\alpha$.

\section{Experimental Results}

\begin{figure*}
    \centering
    \includegraphics[width=1\linewidth]{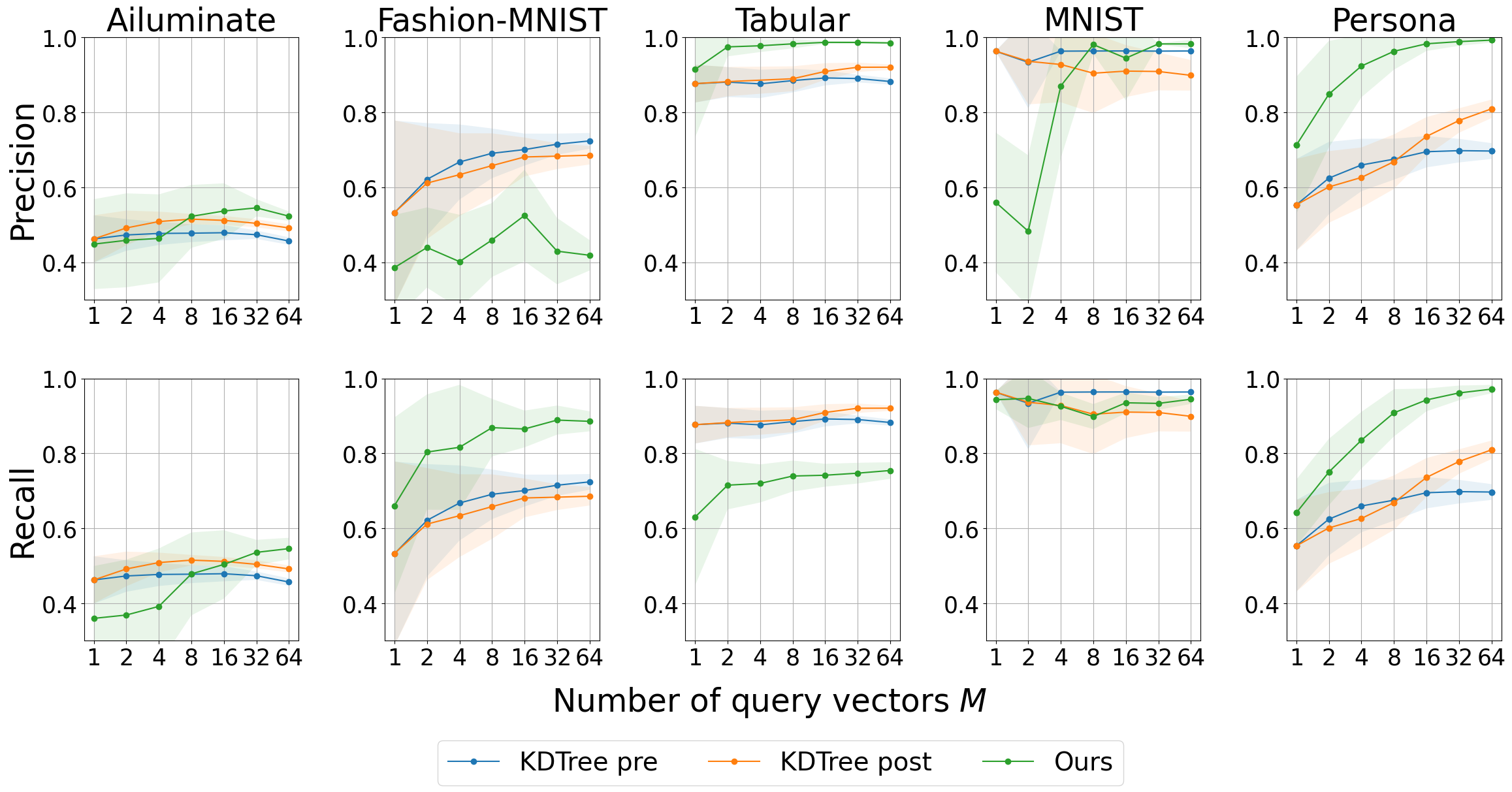}
    \caption{Precision ($\uparrow$) and recall ($\uparrow$) for varying number of query vectors $M=|Q|$ in the query set. Mean (line) and std (shaded) over $50$ random draws of query vectors in $Q$.}
    \label{fig:precision}
\end{figure*}
\subsection{Data}
We evaluate the effectiveness of our retrieval algorithm on two image datasets, \mnist~\cite{lecun2002gradient} and \fashionmnist~\cite{xiao2017fashion}, two textual datasets namely persona~\cite{ perez2023discovering} and Ailuminate ~\cite{ghosh2025ailuminate}, and a tabular dataset from the medical domain\footnote{\label{tabular}\url{https://www.kaggle.com/code/ossm03/knn-algorithm} {, last accessed on 02 Nov 2025.}}. More details in \Appendix~\ref{apx:data}.
For all datasets, we randomly draw vectors $\queryset$ from the set of vectors belonging to a particular class of interest $\classset$; vectors not in $\queryset$ constitute the database $\database$.

\paragraph{(Fashion-)\mnist.}
\mnist\ (handwritten digits) and \fashionmnist\ (Zalando article images) are widely used image datasets, each containing $60{,}000$ grayscale images of size $28 \times 28 \times 1$.\footnote{We use the Torchvision datasets: \url{https://docs.pytorch.org/vision/0.8/datasets.html}, {last accessed on 02 Nov 2025.}} 
We flatten each into a vector in $\mathbb{R}^{784}$, which we then further projected into a lower-dimensional latent space to mitigate sparsity. To capture non-linear structure and preserve local manifold geometry, we apply Uniform Manifold Approximation and Projection (UMAP)~\cite{mcinnes2018umap}. The resulting embeddings have dimensionality $\vectdim= 30$.  
Queries $\queryset$ are drawn from class $2$ (\mnist) and from the \emph{T-shirt/top} class (\fashionmnist), each containing $|\classset| = 6000$ samples.

\paragraph{\persona.}
We create a \persona\ dataset based on a collection of model-generated text datasets describing human values (e.g., political liberalism) associated with different personas~\cite{perez2023discovering}. 
Each dataset corresponds to a single persona and contains sentences labeled as either agreeing or not agreeing with that persona.  
We select $5$ persona datasets for our experiments (see Appendix~\ref{apx:data}). For each persona, we randomly sample $300$ sentences labeled as agreeing with that persona, resulting in a combined dataset of $1{,}500$ sentences.  
We then obtain for each sentence last token embeddings using \emph{LLaMA-3 8B Instruct}~\cite{grattafiori2024llama} with embedding vector dimensionality $\vectdim\ = 4096$. Further details on the embeddings extraction process are provided in \cite{cintas2025localizing}.
Queries $\queryset$ are drawn from the \emph{anti-immigration} persona, with class size $|\classset| = 300$.

\paragraph{\ailuminate.} Ailuminate contains $1{,}200$ human-generated prompts covering $12$ sub-hazard categories. These categories are further grouped into three super-categories: \emph{Physical}, \emph{Non-Physical}, and \emph{Contextual} hazards. Similar to the Persona dataset setup, for each prompt, we extract the last-token embeddings using \emph{LLaMA-3 8B Instruct}. We perform retrieval at the super-category level due to the limited number of samples per sub-hazard category. The query set $\queryset$ is drawn from the \emph{Physical hazard} super-category, with $|\classset| = 500$.

\paragraph{\tabulardata.} We use a tabular dataset from Kaggle\footnote{\label{tabular}\url{https://www.kaggle.com/code/ossm03/knn-algorithm},  {last accessed 02 Nov 2025.}} from the medical domain (breast cancer) with $31$ numerical features.
The dataset consists of $569$ samples and two classes, \emph{Malignant} and \emph{Benign}. Queries $Q$ are drawn from class \emph{Benign} with size $|\classset|=357$.

\subsection{Metrics}
After retrieving a subset of records from the vector database, we assume access to the query class labels for evaluation.  
Positive records are defined as those sharing the same label as the query records, i.e., belonging to $C$, and we compute precision and recall accordingly.
\emph{Precision} is the proportion of correctly retrieved positive records among all retrieved records, reflecting the algorithm's ability to minimize false positives; \emph{recall} is the proportion of correctly retrieved positive records among all positive records in the vector database, reflecting the algorithm's ability to minimize false negatives.\footnote{See \url{https://scikit-learn.org/stable/auto_examples/model_selection/plot_precision_recall.html}, {last accessed  on 02 Nov 2025}.}

\subsection{Baselines}
We compare our retrieval approach with \KDTree~\cite{cunningham2020k}, an algorithm based on Euclidean nearest-neighbour search~\cite{bentley1975multidimensional}, commonly used to efficiently identify similar vectors from a vector database for a given query vector~\cite{cheng2025survey, zolkepli2024multi}. 
\KDTree\ operates on a single query vector and requires the number of vectors to retrieve, $K$, as a input parameter.
 
For any subset of query vectors $\queryset \subset \classset$ drawn from all vectors of a class $C$, we fix the retrieved set size for both methods to the remaining class samples, i.e., $K = |\classset| - |\queryset|$.  
This choice ensures that a recall of 1.0 is achievable. However, it also causes the number of false positives to equal the number of false negatives, leading to identical precision and recall values as observed in \Figure~\ref{fig:precision}.

To extend \KDTree\ to the multi-vector query setting, i.e., when $|\queryset| > 1$, we propose two \KDTree-based baseline variants:  
(i) \KDTpre, which, as a preprocessing step, averages all query vectors in $\queryset$ element-wise across dimensions to form a single average query vector, which is then passed to the standard \KDTree\ algorithm; and  
(ii) \KDTpost, which applies the \KDTree\ algorithm independently to each query vector in $\queryset$, retrieving $K$ vectors per query, and then, as a postprocessing step, selects the top-$K$ vectors ranked by Euclidean distance from the set formed by the union of all retrieved vectors.

\subsection{Results}
\Figure~\ref{fig:precision} reports the mean and standard deviation of precision and recall over $50$ runs for different query set sizes $M = |\queryset|$. In each run, a different random query set $\queryset$ is drawn from the set of vectors $\classset$ belonging to a class of interest. The records retrieved from $\database$ given $\queryset$ are expected to belong to the same class, i.e., $\classset$.

Overall, we observe that across datasets (with the exception being \mnist), larger query set sizes $\numqueries$, leads to higher precision and recall for ours as well as the baseline methods. 
The improvement is most pronounced when increasing $\numqueries$ from $1$ to $8$, while the gains become smaller beyond that (e.g., from $8$ to $16$). This trend is particularly visible in the \persona\ and \tabulardata\ datasets.  This suggests that adding more queries eventually yields diminishing returns. For \KDTpost, moderate improvements continue with larger $\numqueries$.

Analyzing individual datasets, we find that for \mnist\ there is no significant difference between our method and the baselines; however, both precision and recall are relatively high in absolute terms (\~0.9-1.0).
For \fashionmnist, the baselines achieve precision and recall of $\approx0.70$ for $\numqueries=64$, whereas our method reports lower precision but higher recall across all $\numqueries$.
We observe a similar pattern for the \tabulardata\ dataset, where the baselines reach around $0.90$ for both precision and recall, while our method attains a higher precision (almost $1.00$) but a lower recall (about $0.77$). The best performance of our method is observed for the \persona\ dataset, where for $\numqueries=64$, we achieve both precision and recall close to $1.00$, outperforming the baselines, which reach approximately $0.70$ for \KDTpre and $0.80$ for \KDTpost.

In summary, our method consistently retrieves relevant query vectors, often with high precision and recall, whereby the performance varies across datasets.

These results should be interpreted in light of the fact that \KDTree\ retrieves a fixed,  large, number of database vectors $K$, whereas our method dynamically determines the retrieval set size. In Appendix~\ref{ablation}, we present an ablation comparing different \KDTree\ retrieval sizes alongside the number of vectors retrieved by our method. Specifically, our method retrieves approximately $2K$ vectors for \fashionmnist, roughly $K$ for \mnist\ and \persona, and about $2K/3$ for \tabulardata.
When \KDTree\ retrieves a similar number of records as our method, we still slightly outperform it in both precision and recall for the \tabulardata\ and \persona\ datasets.

\section{Limitations and Outlook}

We have presented a novel retrieval approach based on anomalous pattern detection for multi query vector retrieval. Our method represents a first step toward robust multi-query retrieval. While preliminary results on four diverse datasets successfully validate the approach, our study has several limitations.

First, our experiments used fixed classes of interest per dataset. In future work, we plan to explore whether the results generalize across different dataset classes.  

Second, we rely on \KDTree\ as a simple yet robust baseline. In future work, we will incorporate more sophisticated retrieval methods {such as optimized vector similarity search}~\cite{douze2025faiss}.  

Third, we observed that our method performs best on LLM text embeddings compared to reduced-dimensional image embeddings and tabular data. Future work will investigate whether this generalizes to other datasets and holds across different types of LLM embeddings.

\bibliography{refs}

\appendix

\section{Additional Information on DeepScan}
\label{apx:deepscan}
Deep subset scanning (DeepScan) \cite{cintas2021detecting} has been employed in prior work to analyze data for detecting anomalous samples across various computer vision and audio tasks. Applications include, but are not limited to, characterizing the creativity of generative models \cite{cintas2022towards}, detecting adversarial attacks in the inner layers of autoencoders using audio signals \cite{akinwande2020identifying}, classifying skin conditions \cite{kim2022out} and detecting hallucinations in large language model's activations (LLMs) \cite{rateike2023weakly}. 

In the aforementioned prior work, anomalousness is quantified by a scoring function $F(S)$, and the objective is to find the optimal subset of node activations\footnote{A node activation is the value produced by a specific neuron in an embedding vector.} $S^{\star}$, that maximizes this function: $S^{\star} = \arg\max_{S}F(S)$. A suitable choice for the scoring function is the non-parametric scan statistic (NPSS), as adopted in previous studies \cite{cintas2021detecting,cintas2022towards,akinwande2020identifying,kim2022out,rateike2023weakly}.

The general form of the NPSS scoring function is defined as:
\begin{equation}
    F(S)=max_{\alpha}F_{\alpha}(S)=max_{\alpha}\phi(\alpha,N_{\alpha}(S),N(S)).
\end{equation}
Here, $N(S)$ denotes the number of $p$-values contained in the subset $S$, and $N_{\alpha}(S)$ represents the number of $p$-values within $S$ that are less than $\alpha$. Several NPSS scoring functions have been proposed in the literature, including the Kolmogorov–Smirnov test \cite{kolmogorov1933sulla}, the Higher Criticism test \cite{donoho2004higher}, and the Berk–Jones test statistic \cite{berk1979goodness}. 

In this work we use the Berk-Jones test statistic $\phi_{BJ}$~\cite{berk1979goodness}, which is defined as:
\begin{equation}
   \phi_{BJ}(\alpha,N_{\alpha},N)=N\times KL\left(\frac{N_{\alpha}}{N},\alpha\right). 
\end{equation}
Where $KL$ represents the Kullback-Leibler divergence~\cite{kullback1951kullback}, between the observed and expected proportions of $p$-values.

Maximizing the scoring function over all possible subsets of dimensions and vector data points is computationally challenging. FGSS approaches this problem by reducing the number of subsets under consideration from $O(2^{M})$ to a linear search of $O(M)$, where $M$ denotes the number of elements currently being optimized (such as dimensions or vector data points). FGSS has been shown to be effective in prior work \cite{kim2022out,cintas2021detecting}. Its computational efficiency is guaranteed by the use of Linear-Time Subset Scanning (LTSS) \cite{neill2012fast}. LTSS operates by sorting elements according to a priority score, defined as the proportion of $p$-values below a specified threshold $\alpha$. Ultimately, FGSS returns the subset of nodes $O_{S}^{\star}$ and the corresponding subset of vector data points from the dataset.

\section{Data Preprocessing}\label{apx:data}

\subsection{Persona Dataset}

When performing retrieval on the entire Persona dataset using \emph{anti-immigration} as the candidate query class, we observed considerably low precision and recall. This occurred because a large number of records were retrieved, many of which were false positives. Consequently, we discarded classes with the highest number of False positive counts. We ended up selecting \textit{conscientiousness},~\textit{neuroticism},~\textit{openness},~\textit{anti-immigration} and ~\textit{subscribes-to-deontology} for our experiments.

\subsection{Tabular Dataset}
 We use a tabular dataset from Kaggle\footnote{\label{tabular}\url{https://www.kaggle.com/code/ossm03/knn-algorithm}, {last accessed on 02 Nov 2025}.} from the medical domain (breast cancer) with $31$ numerical features. The dataset
contains different measurements and characteristics of cell nuclei present in breast cancer biopsies,
 such as the radius of the cell nucleus, is concavity, or its smoothness.\footnote{For more information, refer to~\url{https://www.kaggle.com/datasets/gkalpolukcu/knn-algorithm-dataset/discussion/468599}, {last accessed on 02 Nov 2025}.}
\section{Additional Results}\label{ablation}
\begin{figure*}
    \centering
    \includegraphics[width=\linewidth]{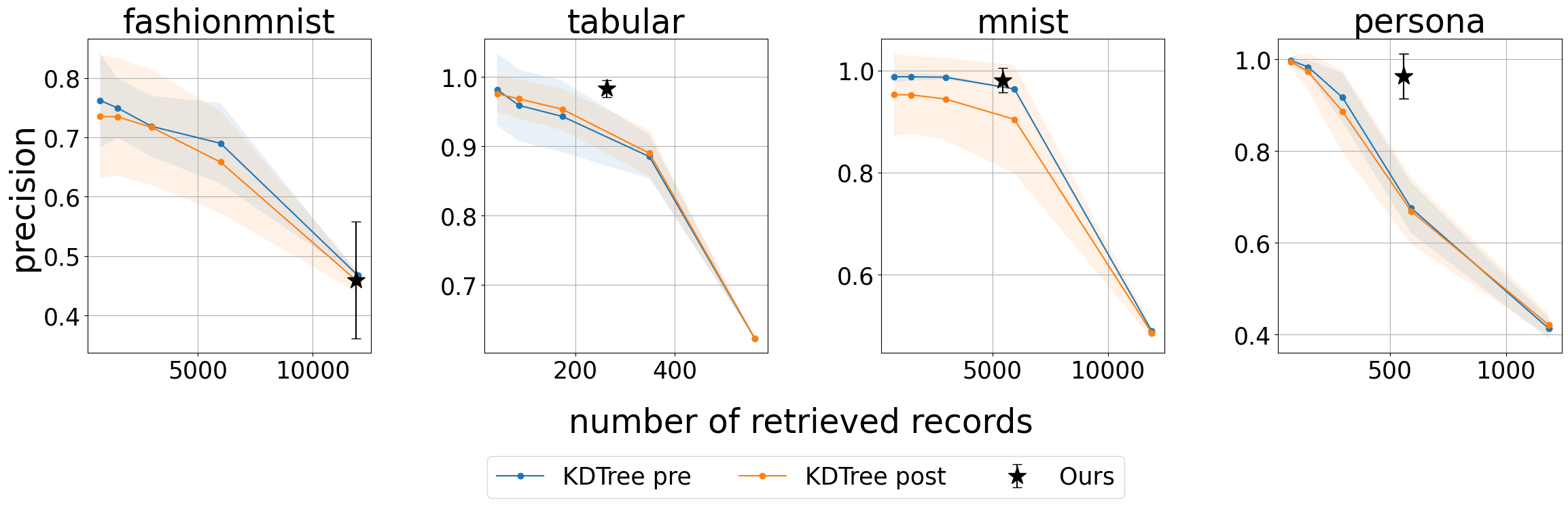}
    \caption{As the number of records to be retrieved ($K$) increases, the precision ($\uparrow$) of \emph{KDTree pre} and \emph{KDTree post} decreases. In contrast, our approach maintains consistent precision, since $k$ is determined automatically rather than being a tunable parameter.}
    \label{fig:ablation-precision}
\end{figure*}

\begin{figure*}
    \centering
    \includegraphics[width=\linewidth]{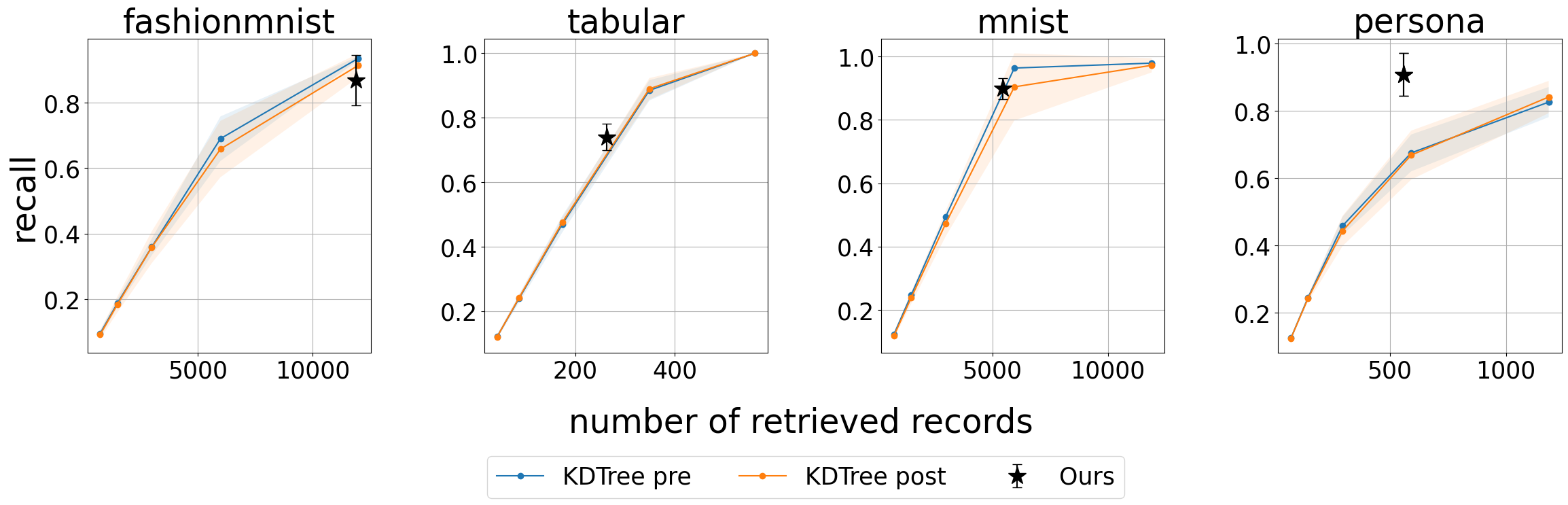}
    \caption{As the number of records to be retrieved ($K$) increases, the recall ($\uparrow$) of \emph{KDTree pre} and \emph{KDTree post} increase. It is important to note here that, $k$ is set to values even higher than the number of samples of the candidate class. }
    \label{fig:ablation-recall}
\end{figure*}

We present additional results analyzing the influence of the hyperparameter $K$ on the performance of the baselines. In the main paper results, we fixed the number of retrieved records $K$ in \KDTpre\ and \KDTpost. Here, we systematically vary $K$ and, for comparison, report the corresponding number of retrieved queries for our method. Precision results are shown in \Figure~\ref{fig:ablation-precision}, and recall results in \Figure~\ref{fig:ablation-recall}.

\end{document}